\documentclass[runningheads]{llncs}

\usepackage{eccv}

\usepackage{eccvabbrv}

\usepackage{graphicx}
\usepackage{booktabs}
\usepackage{adjustbox}
\usepackage{multirow}
\usepackage[accsupp]{axessibility}  % Improves PDF readability for those with disabilities.

\usepackage{hyperref}

\usepackage{orcidlink}
\newcommand{\bftab}{\fontseries{b}\selectfont}
\newcommand{\fakeparagraph}[1]{\noindent{\bftab #1}}

\begin{document}

\title{A Simple Background Augmentation Method for Object Detection with Diffusion Model} 

\titlerunning{Background Augmentation with Diffusion Model}

\author{Yuhang Li\inst{1}\orcidlink{0000-0002-6444-7253}\and
Xin Dong\inst{1} \and Chen Chen\inst{1} \and Weiming Zhuang\inst{1} \and Lingjuan Lyu\inst{1}\thanks{Corresponding Author. Work done during Yuhang Li’s internship at Sony AI.}}

\authorrunning{Y. Li, X. Dong, C. Chen, W. Zhuang, L. Lyu.}

\institute{Sony AI\\
\email{\{yuhang.li, xin.dong, chen.chen, weiming.zhuang, lingjuan.lyu\}@sony.com}\\
}

\maketitle

\begin{abstract}
In computer vision, it is well-known that a lack of data diversity will impair model performance.
In this study, we address the challenges of enhancing the dataset diversity problem in order to benefit various downstream tasks such as object detection and instance segmentation. We propose a simple yet effective data augmentation approach by leveraging advancements in generative models, specifically text-to-image synthesis technologies like Stable Diffusion. Our method focuses on generating variations of labeled real images, utilizing generative object and background augmentation via inpainting to augment existing training data without the need for additional annotations. We find that background augmentation, in particular, significantly improves the models' robustness and generalization capabilities. 
We also investigate how to adjust the prompt and mask to ensure the generated content %do not violate
comply with the existing annotations. 
The efficacy of our augmentation techniques is validated through comprehensive evaluations of the COCO dataset and several other key object detection benchmarks, demonstrating notable enhancements in model performance across diverse scenarios. This approach offers a promising solution to the challenges of dataset enhancement, contributing to the development of more accurate and robust computer vision models.
\end{abstract}

\section{Introduction}

The quest for robust and accurate object detection~\cite{he2017mask, redmon2016you, feng2021tood, feng2022promptdet } and instance segmentation~\cite{hafiz2020survey} models is a cornerstone for contemporary computer vision research. These models underpin a variety of applications, from autonomous vehicles~\cite{hnewa2020object} to medical image analysis~\cite{litjens2017survey, yang2021artificial}, necessitating high levels of precision and reliability. Central to the development of these models is the availability of large, diverse, and accurately annotated datasets. However, creating such datasets is fraught with challenges, including the high costs of data collection and annotation, privacy concerns, and the potential for dataset bias. For example, collecting images of urban driving scenes requires physical car infrastructure, and labeling a segmentation annotation for a single urban image in Cityscapescan takes up to 60 minutes ~\cite{cordts2016cityscapes}. 
Therefore, it is necessary to navigate new solutions to address these limitations, \eg data augmentations and low-cost annotations. 
 
Recent advancements in generative models, particularly in the domain of text-to-image synthesis~\cite{nichol2021glide, ramesh2022hierarchical, saharia2022photorealistic} (e.g., Stable Diffusion~\cite{rombach2022high}), offer a promising avenue for addressing aforementioned challenges. %of data augmentation.
While Stable Diffusion and similar models can create visually compelling images, generating synthetic images that are accompanied by accurate and detailed object annotations remains a formidable challenge. This complexity stems from the need for precise alignment between the generated visual content and its corresponding annotations. Although several works have %implemented some solutions to 
generated the synthetic data and instances annotations simultaneously~\cite{wu2023diffumask, yang2024freemask, wu2024datasetdm}, they all require finetuning the existing text-to-image on the current dataset and generate a significant amount of synthetic data for pre-training, which increases training complexity significantly.

To this end, we explore generating variants of labeled real images for object detection and instance segmentation while reusing their annotations, in a way that augments the existing training data. Specifically, we explore generative object augmentation and background augmentation through inpainting, which allows for the enhancement of objects and backgrounds respectively within the dataset. 
Notably, we discover that background augmentation boosts the model performance, which suggests that modifying the 
% our findings indicate a superior performance of background augmentation, suggesting that modifications to the 
background context of images plays a crucial role in improving model robustness and generalization.

We rigorously evaluate our augmentation methods on the MS COCO dataset \cite{lin2014microsoft} and PASCAL VOC \cite{everingham2010pascal} for both detection and segmentation tasks across multiple architectures. This comprehensive evaluation demonstrates the effectiveness of our proposed augmentation techniques in improving the performance of object detection and instance segmentation models under various challenging scenarios. We summarize our contributions as follows:
\begin{enumerate}
    \item We introduce a simple data augmentation framework for object detection with the text-to-image model without finetuning. We find that augmenting the background brings more advantages than augmenting objects. 
    \item We carefully craft the text prompt, mask region, and diffusion steps to ensure the generated content is effective for data augmentation purposes. The ablation studies show that they are essential to achieve superior performance.
    \item We evaluate the background augmentation policies on various tasks including object detection and instance segmentation, significantly improving their performance. For example, our background augmentation increases mAP up to 5.3\% when using only 10\% of the COCO training data. 
    
\end{enumerate}

\section{Related Work}

This section provides literature reviews of object detection, data augmentation, and text-to-image generation for visual tasks.

\fakeparagraph{Object Detection.} 
Object detection aims to simultaneously predict the category and corresponding bounding box
for the objects in the images. Generally, object detectors~\cite{redmon2016you, ren2015faster, he2017mask} are trained on a substantial amount of
training data with bounding box annotations and can only
recognize a predetermined set of categories present in the
training data, \eg MS COCO dataset~\cite{lin2014microsoft}. 
Both backbone architecture and the detection neck/head framework are important for object detection performance. For backbone, Vision Transformer~\cite{vit, liu2021swin, li2022exploring, he2021masked} has emerged as an effective backbone alternative compared to convolutional neural networks~\cite{he2016deep}. These transformer-based models leverage self-attention mechanisms to capture global dependencies within the image. 
Moreover, the development of efficient and scalable detectors, such as EfficientDet~\cite{tan2020efficientdet}, further underscores the importance of optimizing both accuracy and computational efficiency.

\fakeparagraph{Data Augmentation. }
Data augmentation~\cite{shorten2019survey,li2024synthetic} plays an indispensable role in the deep learning model, as it forces the model to learn invariant features, and thus helps generalization.
The data augmentation is applied in many areas of vision tasks including object recognition~\cite{cubuk2018autoaugment,lim2019fastautoaug,cubuk2020randaugment}, video perception~\cite{yun2020videomix}, and semantic segmentation~\cite{ronneberger2015u}. 
Apart from learning invariant features, data augmentation also has other specific applications in deep learning. For example, adversarial training~\cite{ganin2016domain,tramer2017ensemble} leverages data augmentation to create adversarial samples and thereby improves the adversarial robustness of the model.
In terms of data augmentation in object detection, Zoph et.al \cite{zoph2020learning} propose reinforcement learning to search for complex augmentation policies including color transformations, geometric transformations, and bounding box transformations. CutMix~\cite{yun2019cutmix} proposes a spatial mixing strategy for augmenting the detection results. RandAugment~\cite{cubuk2020randaugment} uses a reduced search space for searching both classification and detection augmentation policies.

\fakeparagraph{Text-to-Image Data for Visual Tasks. }
Traditional synthetic data are acquired through the renderings from the graphics engines~\cite{dosovitskiy2015flownet, peng2017visda, richter2016playing}. However, this type of synthesis cannot guarantee the quality and diversity of the generated data, resulting in a large gap with real-world data. Recent success in generative models has made synthesizing photo-realistic and high-fidelity images possible, which could be used for training the neural networks for image recognition due to their unlimited generation. 
For example, early works explored Generative Adversarial Networks~(GAN)~\cite{creswell2018generative} for image recognition tasks. \cite{besnier2020dataset} uses a class-conditional GAN to train the classifier head and \cite{zhang2021datasetgan} uses StyleGAN to generate the labels for object segmentation. \cite{jahanian2021generative} adopts the GAN as a generator to synthesize multiple views to conduct contrastive learning. 
Until recently, the text-to-images models~\cite{dhariwal2021diffusion, lugmayr2022repaint, rombach2022high, saharia2022palette} have been leveraged to synthesize high-quality data for neural network training due to their effectiveness and efficiency. 
\cite{he2022synthetic} adopts GLIDE~\cite{nichol2021glide} to synthesize images for classifier tuning on the CLIP model~\cite{radford2021learning}. However, simply tuning the classifier may not fully explore the potential of synthetic data. StableRep~\cite{tian2023stablerep} proposes to use Stable Diffusion for generating pre-trained datasets for contrastive learning and leverages the synthetic data from different random seeds as the positive pairs. 
\cite{azizi2023synthetic} explores synthesizing data under ImageNet label space and %finds
produces improved performance. Fill-up~\cite{shin2023fill} balances the long-tail distribution of the training data using the synthetic data from text-to-image models.
\cite{wu2023diffumask, wu2024datasetdm} propose to leverage text-to-image models for segmentation tasks where they adjust the model to generate both masks and images. However, this type of method requires either finetuning the model on the target dataset or synthesizing a large amount of pre-trained data. 

\section{Generative Background Augmentation}

In this section, we introduce our generative background augmentation method using the text-to-image diffusion model to increase the diversity of the original dataset. We first discuss the principle of the Inpainting technique and then adapt this method to our background augmentation framework. 

\subsection{Generative Augmentation with Inpainting}

Our primary data synthesis model is Stable Diffusion~\cite{rombach2022high}. 
Formally, 
% denote the text language prompt embedding as $\mathbf{s}$,
the Stable Diffusion generates images from a random Gaussian noise $\mathbf{z}\sim\mathcal{N}(\mathbf{0, 1})$ by applying the denoising model $\epsilon(\cdot)$ repeatedly. We denote the text language prompt embedding as $\mathbf{s}$. In this work, we apply the widely-used classifier-free guidance~\cite{ho2022classifier} which linearly combines the conditional estimate and the unconditional estimate of noise, given by
\begin{equation}
\label{equ:}
    \tilde{\epsilon}(\mathbf{s, z}_t) = w\epsilon(\mathbf{s, z}_t) + (1-w) \epsilon(\mathbf{z}_t),
\end{equation}
where $w$ is the guidance scale and $t$ is the step of the denoising process. After multiple denoising steps, Stable Diffusion employs a decoder $\mathcal{D}$ to decode the latent to the image. 

Given an input image $\mathbf{x}$, the inpainting task is to re-generate selective areas in the image indicated by a mask variable $\mathbf{m}$ while keeping other pixels unchanged. Since Stable Diffusion conducts the generation process in the latent space, we consider inpainting in the latent space as well. First, the encoder projects input image to latent space $\hat{\mathbf{x}} = \mathcal{E}(\mathbf{x})$. Similarly, the mask variable is also projected to the latent space by resizing it $\hat{\mathbf{m}}=\text{resize}(\mathbf{m})$ to have the same size of $\hat{\mathbf{x}}$. 
% To perform inpainting for a given image, a mask variable $\mathbf{m}$ is provided to indicate where to inpaint the desired object or background from the text prompt and an original image variable $\mathbf{x}$ is used to replace unmasked region. First, the image variable is encoded by the encoder $\hat{\mathbf{x}} = \mathcal{E}(\mathbf{x})$, then the diffusion process (essentially modeled by a Markov chain)  $q(\hat{\mathbf{x}}_t|\hat{\mathbf{x}}_{t-1}) = \mathcal{N}(\hat{\mathbf{x}}_t, \sqrt{1-\beta_t}\hat{\mathbf{x}}_t, \beta_t\mathbf{I}))$ will add noise to the image latent at $t$-th step, where $t\in[1, T]$. The latent noise at the final timestep $T$ is equivalent to the random Gaussian noise $\mathbf{z}$.
Now, the inpainting denoising step can be formulated as 
\begin{equation}
    \hat{\epsilon}(\mathbf{s}, \hat{\mathbf{x}}_t) = \hat{\mathbf{m}}\odot\tilde{\epsilon}(\mathbf{s}, \hat{\mathbf{x}}_t) + (\mathbf{1}-\hat{\mathbf{m}})\odot(\hat{\mathbf{x}}_{t} - \hat{\mathbf{x}}_{t-1}). 
\end{equation}
Here, $\odot$ denotes the element-wise multiplication operation. 
% As a result, the Inpainting only changes the masked region of the original image $\mathbf{x}$ and aligns the content with the text prompt. 
% Therefore, Inpainting is a suitable off-the-shelf tool to keep the annotation intact as it can specify the region. 
The inpainting process described above modifies only the masked region, denoted as $\hat{\mathbf{m}}$, within the original image's latent space. Owing to the localized nature of Stable Diffusion's decoding mechanism, this method can also provide a reasonable assurance that solely the region specified by $\mathbf{m}$ will be regenerated within the image space.

\subsection{Object Augmentation or Background Augmentation?}

Data augmentation serves the role of increasing data amount and diversity to optimize for generalization performance. Traditionally, data augmentation in object detection is conducted through image transformations. For example, Zoph et.al~\cite{zoph2020learning} define three kinds of data augmentation: (1) color operations, (2) geometric transformations, and (3) bounding box operations, which change the image illumination, position, and annotation respectively. However, these augmentation methods only change pixel values and keep the content of the object or background the same as the original images.
% potentially lacks diversity as the object or the background is kept the same. 

In this work, we explore another kind of augmentation method, \ie using generative models like diffusion models to change the content of training images to increase the training data volume and diversity. 
Specifically, consider a training image $\mathbf{x}\in\mathbb{R}^{3\times H\times W}$ containing $n$ different objects, where each object corresponds to a unique mask variable $\mathbf{m}_i\in\{0, 1\}^{H \times W}$. The background mask can be determined by the following formula, 
\begin{equation}
    \mathbf{m}_b = \mathbf{1} - \sum_{i=1}^n \mathbf{m}_i. 
\end{equation}
With different choices from $\{\mathbf{m}_b, \mathbf{m}_i\}$, we are able to change the content of different regions of the original image. 

Specifically for dense vision tasks like object detection and instance segmentation, a natural question is whether we should augment objects, background, or both of them.
Surprisingly, we found object augmentation is not suitable in practice.
% does not bring significant improvement in practice. 
The reasons are multiple. 

\begin{figure}[t]
    \centering
    \includegraphics[width=\linewidth]{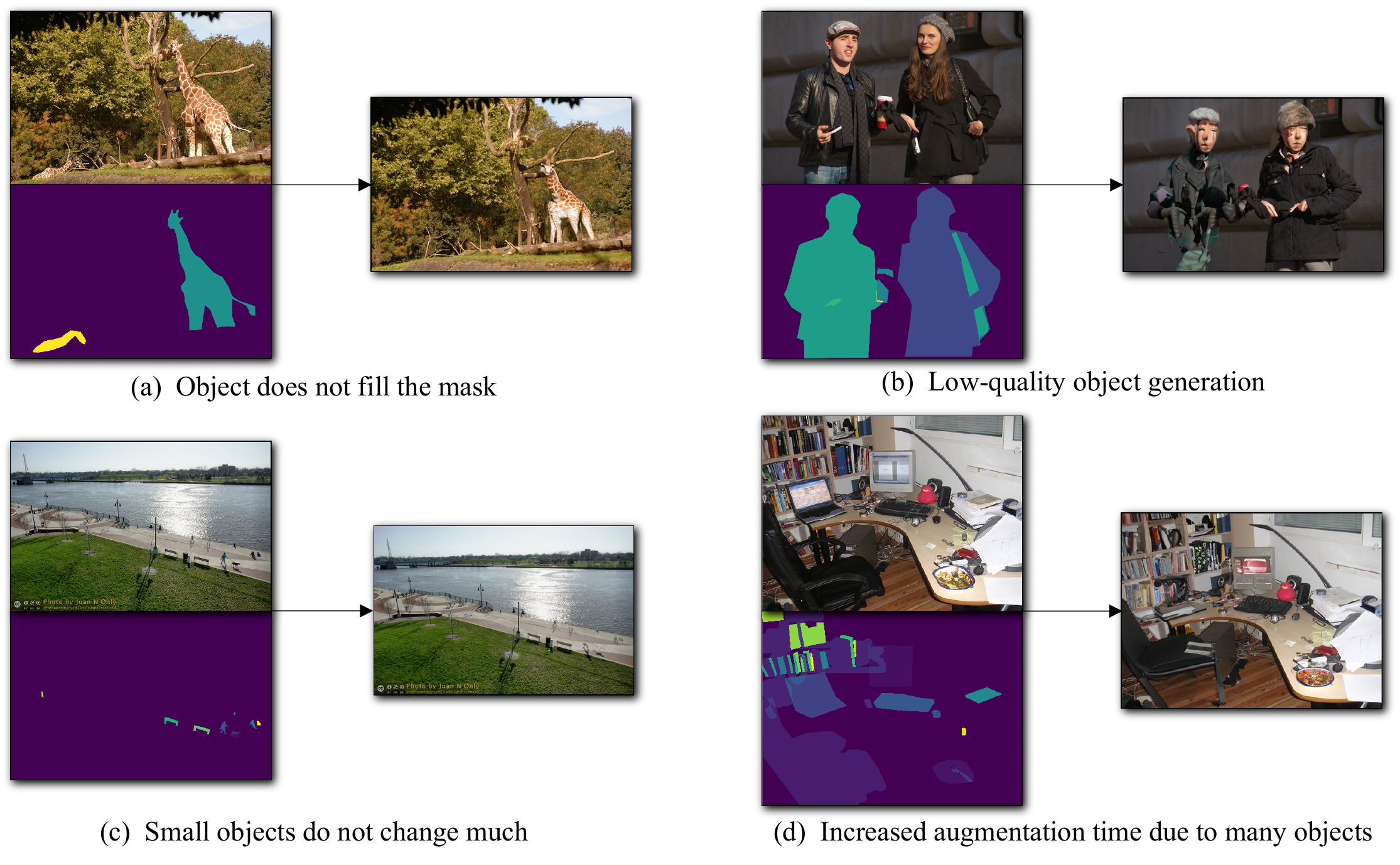}
    \caption{Examples of failed object augmentation in MS COCO dataset using stable diffusion v1-5.}
    \label{fig_object_aug}
\end{figure}

First, generating a realistic object that can precisely fill the mask region is quite challenging for existing Stable Diffusion models. For example, if a generated object cannot adequately cover its corresponding mask region, reusing the mask as its annotation will mislead the training of the model. We visualize some exemplary failure cases of object augmentation in Fig.~\ref{fig_object_aug}(a). Although some efforts have been investigated to improve the alignment between generated objects and their corresponding masks~\cite{xie2023smartbrush, xue2023freestyle}, they either introduce prohibitive computational overhead for large datasets or achieve an unsatisfactory mask-object alignment level, %that is required to enhance
which is harmful for enhancing task performance for object detection and instance segmentation.

In addition, we have observed that object augmentation can yield low-quality inpainting results for certain object classes, especially those involving human figures~\cite{weng2023diffusion}. This can adversely affect the quality of the data and may even degrade the performance of vision-related tasks (as shown in \cref{fig_object_aug}(b)). Furthermore, augmenting small objects results in negligible improvements because they contain relatively few pixels (\cref{fig_object_aug}(c)). Consequently, the detection and segmentation of small objects do not benefit from object augmentation.

Finally, data generation costs of object augmentation is much higher than background augmentation.
While for background augmentation, only one inpainting is enough for the background area. For example, the averaged number of objects in the MS COCO training dataset~\cite{lin2014microsoft} is 7.27, amounting to $\sim$7.3$\times$ generation cost when comparing object augmentation to background augmentation (\cref{fig_object_aug}(d)).

% Second, controlling each object requires filling the mask region with the correct pose and angles. As we show in Fig. X, the filling of object detection cannot be guaranteed.

In the experiments section, we present a thorough ablation study demonstrating that object augmentation underperforms compared to background augmentation and can even lead to a degradation in downstream task performance when compared to a baseline without generative augmentation.

\subsection{Utility-Aware Background Augmentation}

In this section, we concentrate on background augmentation techniques with high training utility. An optimal background augmentation approach must avoid introducing extraneous objects, as these can confound the training process due to the lack of corresponding annotations. Additionally, it is crucial for a background augmentation method to preserve the integrity of existing objects, thereby maintaining a high level of object-mask alignment. We have investigated several design strategies to fulfill these requirements.

\fakeparagraph{Prompt Selection of Background Augmentation. }
A critical choice for data generation is the text prompt $\mathbf{s}$ for guiding the inpainting process. For object augmentation, one can safely adopt the class label as the text prompt while for background augmentation a text description for the whole background is needed. 

Some datasets like MS COCO provide image caption annotations for each image, which could be a potential text prompt for background augmentation. 
However, applying an image caption as a text prompt directly brings two disadvantages in practice. 
First, the image caption usually contains object descriptions as well. Using this caption will generate additional objects in the background. Second, not all detection and segmentation datasets contain image captions. It would require other captioning models to generate the new caption for the image, which involves additional computation costs and data cleaning to get accurate captions. 

To this end, we use a simple text prompt ``\texttt{Generate a clean background}'' for background augmentation. We found this choice is sufficient for Stable Diffusion to modify the background significantly without generating additional objects. 

\begin{figure}[t]
    \centering
    \includegraphics[width=\linewidth]{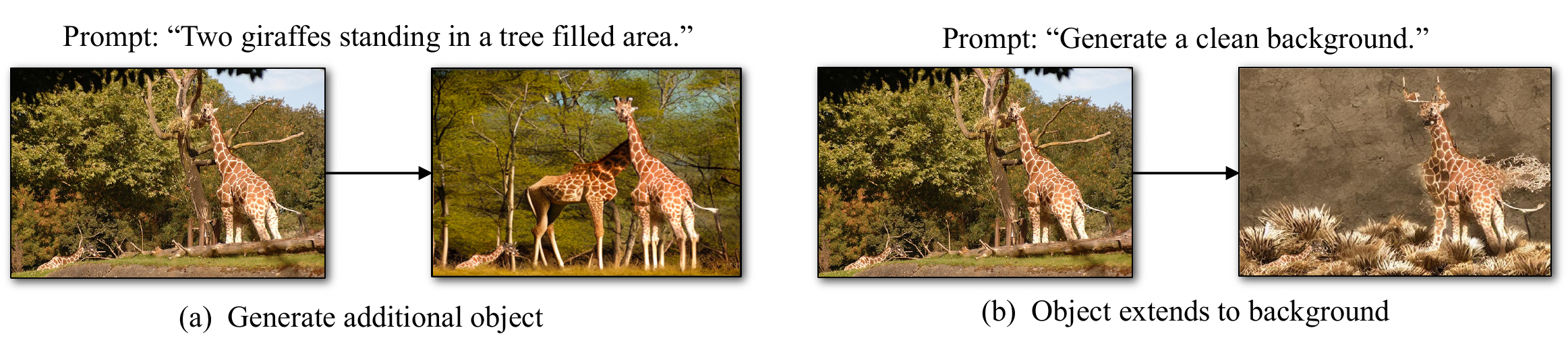}
    \caption{Visualization of several failure cases for background augmentation. (a) using image caption as text prompt, (b) using our prompt method but the objects extend to the background. }
    \label{fig_bg_aug}
\end{figure}

\begin{figure}[t]
    \centering
    \includegraphics[width=\linewidth]{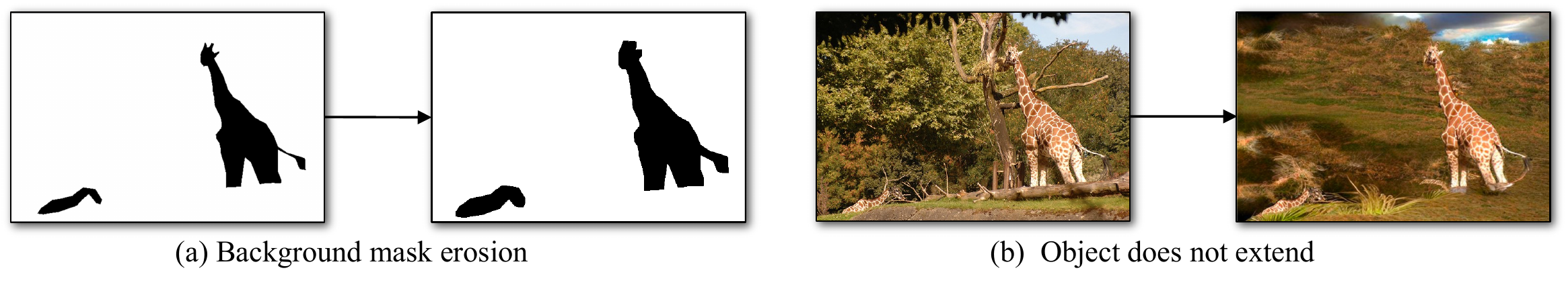}
    \caption{Visualization of our background augmentation: (a) using mask erosion and (b) the results.}
    \label{fig_bg_aug2}
\end{figure}

\fakeparagraph{Background Mask Erosion. }
During the process of background augmentation, we occasionally observed that objects would inadvertently extend into the background region. This phenomenon is illustrated in Fig.~\ref{fig_bg_aug}(b), where the giraffe appears larger than its corresponding mask, thereby disrupting the alignment with its initial annotation. To address this issue, we introduce the concept of background mask erosion. This technique employs a minimum filter to the background mask, specifically, $\mathbf{m}_b=\mathrm{MinimumFilter}(\mathbf{m}_b, k)$. We empirically determined that a kernel size $k$ of 7 is effective in preventing object extension. Fig.~\ref{fig_bg_aug2}(a) demonstrates the process of eroding the background mask with the minimum filter, and Fig.~\ref{fig_bg_aug2}(b) shows a much better result after adopting our background mask erosion.

\fakeparagraph{Adaptive Augmentation Freedom. }
% Given that each image has different object sizes or background sizes, the difficulty of the background augmentation could be different. Images that have a larger background are harder to augment as the degree of freedom is larger. 
Considering the variation in background region sizes across different images, there is an increased likelihood of errors during background augmentation—such as the introduction of undesired objects or the unwarranted extension of objects into the background areas—for images with larger backgrounds, potentially leading to the generation of harmful synthetic data. In the Stable Diffusion inpainting process, the number of diffusion steps can be adjusted to calibrate the extent of modifications applied to the original input image, that is, the distance between the original and the generated image. Drawing inspiration from this aspect of the process, we propose tailoring the degree of background augmentation to the size of the background region.

To this end, we introduce an adaptive control method for the inpainting process. Specifically, we compute the ratio of the background area to the overall image area and utilize this metric to determine the appropriate number of diffusion steps for the original training image $\mathbf{x}$. The relationship is formalized as follows:

\begin{equation}
\hat{T} = T \times \left(1 - D \times \frac{\mathrm{sum}(\mathbf{m}_b)}{W \times H}\right).
\end{equation}

In this equation, $\mathrm{sum}(\cdot)$ denotes the summation of all elements within the input matrix, and $W \times H$ denotes the total area of the training image. 
$D\in(0, 1)$ controls the range of diffusion steps variation.
$T$ is the maximum diffusion step where $\hat{T}=T$ indicates background noise starts from random Guassian noise.  Technically, if the background-to-image area ratio is higher, we reduce the diffusion steps to preserve more information from the original data.

\begin{figure}
    \centering
    \includegraphics[width=\linewidth]{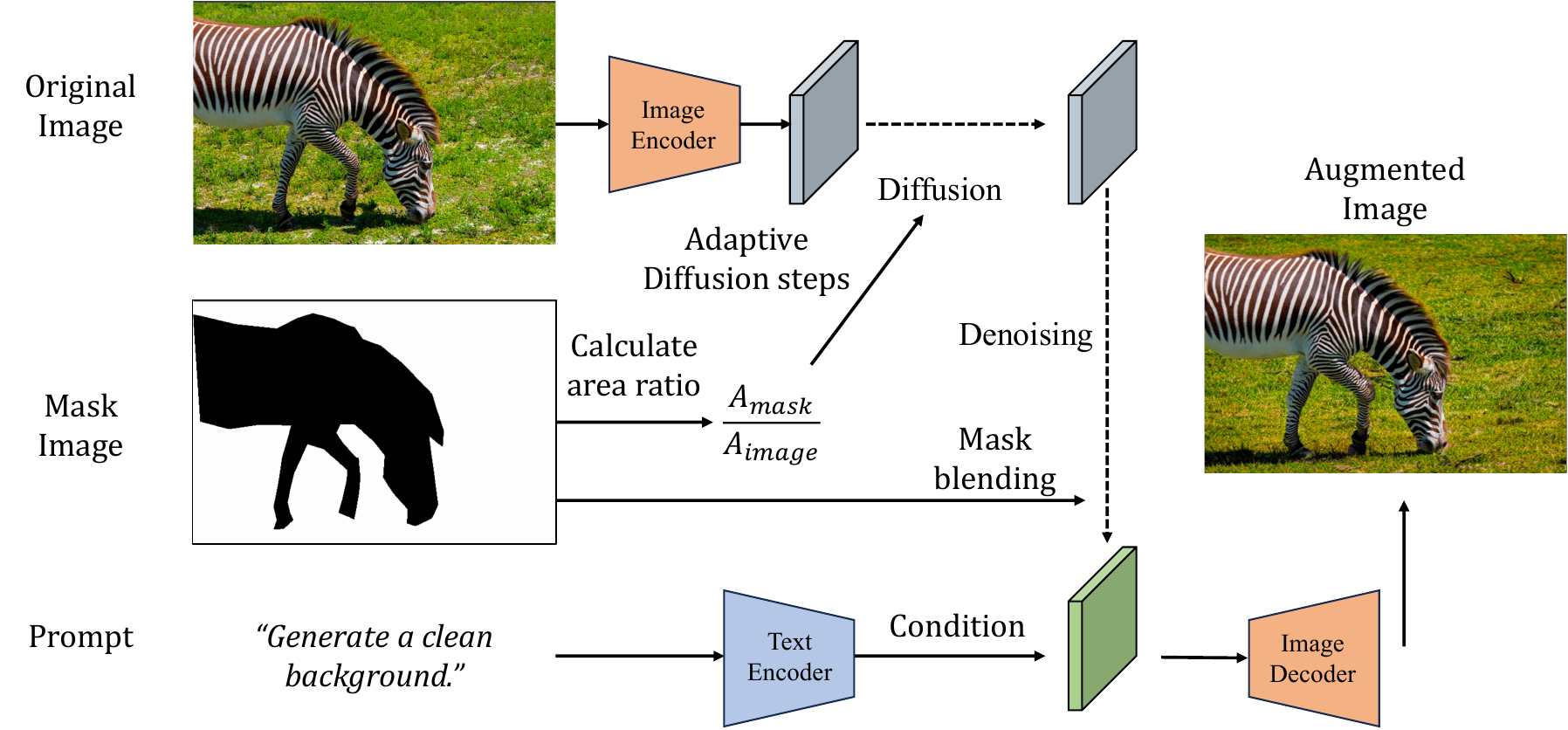}
    \caption{The overall background augmentation framework of our method.}
    \label{fig_framework}
\end{figure}

% \fakeparagraph{Overall Pipeline. }
Fig.~\ref{fig_framework} illustrates the process of our overall background augmentation pipeline. The image is first diffused based on the area ratio calculated from the background mask and then denoised using the eroded mask to generate the new background.

\subsection{Training with Background Augmented Data}
Unlike traditional data augmentation which modifies the image tensor on the fly~\cite{cubuk2020randaugment}, our method generates the additional images and expands the dataset size. Hence, our data augmentation can be combined with conventional data augmentation methods to improve the generalization ability in object detection. 
We also utilize our method with semi-supervised learning. Specifically, we incorporate our augmented data in Soft Teacher~\cite{xu2021end} framework as part of the ``strong augmentation'' to optimize the student network.
The ``weak augmentation'' is applied to the unlabeled images on teacher networks to generate soft annotations for student learning.

\section{Experimental Evaluation}

In this section, we investigate the impact of the augmented data by using our simple background augmentation framework. We first visualize several examples of data augmentation and then report object detection/instance segmentation performance on MS COCO~\cite{lin2014microsoft} and PASCAL VOC~\cite{everingham2010pascal} datasets. 
We also conduct several ablation studies and investigations into different data regimes. 

\begin{figure}[t]
    \centering
    \includegraphics[width=\linewidth]{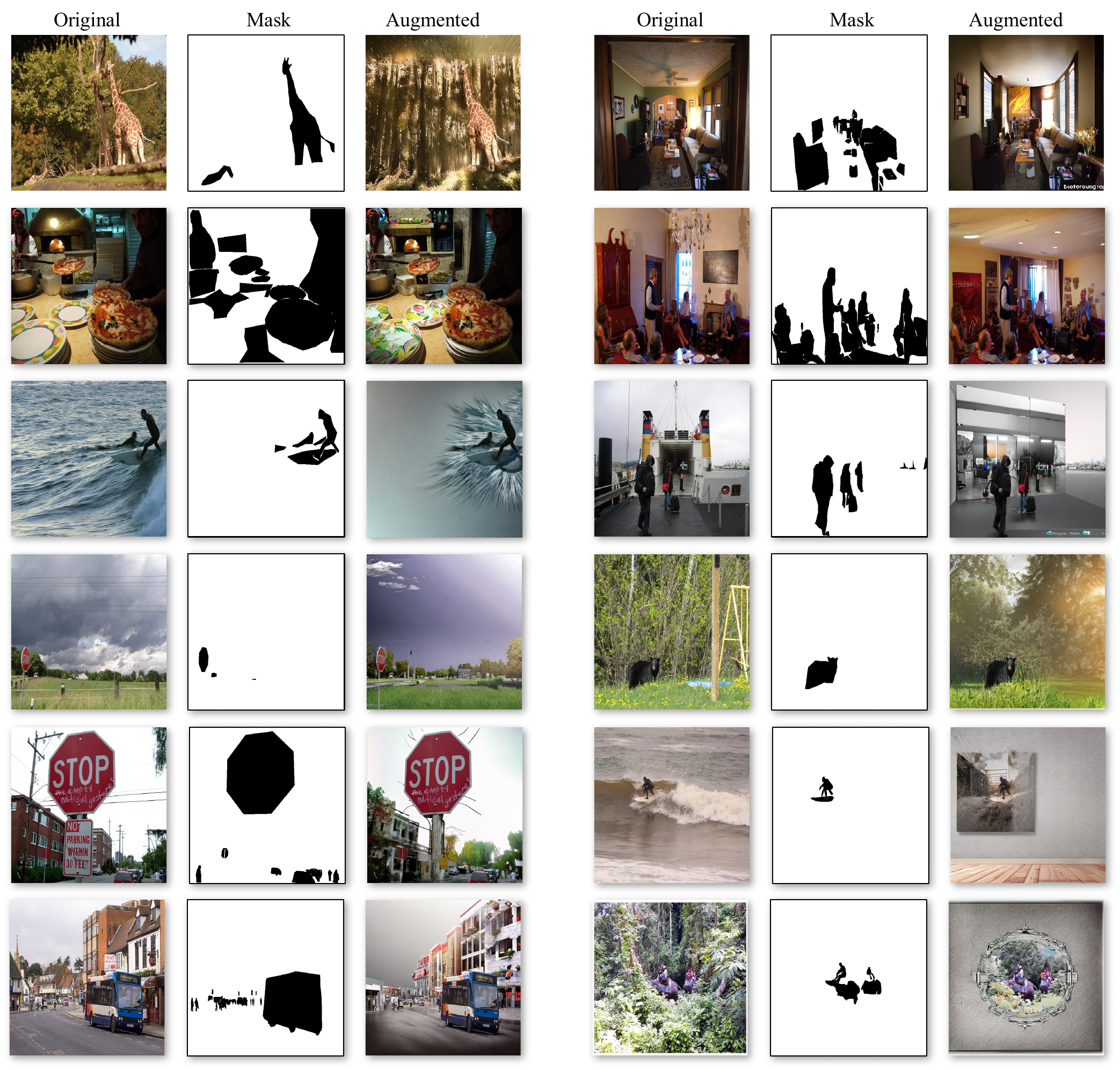}
    \caption{Example images of our background augmentation on MS COCO dataset. }
    \label{fig_examples}
\end{figure}

\subsection{Visualization}
% In order to demonstrate the effectiveness of our background augmentation pipeline for improving object detection data diversity without additional annotation, 
We demonstrate a variety of synthesis results (as well as the original image and the background mask image) using our method on the MS COCO dataset~\cite{lin2014microsoft}, as shown in Fig.~\ref{fig_examples}.
Our approach can modify the background of training data to various extents. In certain instances, the new background remains within a similar scene as the original (e.g., images at the second row and first and second columns), yet exhibits different styles and content. Conversely, some augmented backgrounds may represent entirely new scenes. For example, in the fourth row and first column, the original background of a stop sign is transformed into a purple sky, while maintaining the original stop sign objects unchanged. In the fifth/sixth row second column, our method converts a background into a photo hanging on the wall.

These background augmentations enhance the diversity of the training dataset without introducing additional annotation burdens or compromising the quality of existing annotations.

\subsection{Object Detection Evaluation}
\label{sec_odeval}
We start with the evaluation of our method on the MS COCO %object detection 
dataset. We train a Faster-RCNN~\cite{ren2015faster} with FPN~\cite{lin2017feature}, using the ResNet-50~\cite{he2016deep} backbone. We also provide results of other datasets and model architectures in \cref{sec_pascal}. 

\fakeparagraph{Implementation Details. }
Specifically, we sampled 1\%, 5\%, 10\%, and 25\% of the original training dataset. By default, background augmentation was applied to the selected training images, generating one augmented copy per image. Both the original and the augmented images were utilized during training. For the noise sampling schedule, we employed Stable Diffusion v2-1~\cite{rombach2022high} and DPMSolver~\cite{lu2022dpm, lu2022dpm++}. We resized the image resolution to $512\times512$ for the inpainting process and subsequently restored the images to their original resolution post-inpainting.

For the training of the detection model, we adhered to the implementation and hyperparameters provided by MMDetection~\cite{chen2019mmdetection} and Soft Teacher~\cite{xu2021end}. We utilized anchors spanning 5 scales and 3 aspect ratios. During the training and inference phases, 2k and 1k region proposals were generated respectively with a non-maximum suppression threshold set at 0.7. During each training iteration, a subset of 512 proposals were sampled from the 2k available to serve as box candidates for training the RCNN. The model underwent training over 12 epochs on 8 GPUs, each processing 5 images. We opted for SGD as the training algorithm, initiating the learning rate at 0.01 and reducing it by a factor of 10 after 110k and 160k iterations. The weight decay was set at 0.0001, while the momentum was maintained at 0.9.

\fakeparagraph{Comparison to Baselines. }
We compare our method against a baseline Faster-RCNN model that does not utilize generative background augmentation. Both the baseline model and our approach incorporate conventional data augmentation techniques, such as horizontal flipping with a 50\% probability and multi-scale jittering, where images are randomly resized within the range of 800 to 1333 pixels. Furthermore, we contrasted our generative background augmentation with the state-of-the-art augmentation method, RandAugment~\cite{cubuk2020randaugment}. RandAugment conducts an exhaustive search for complex augmentation policies across 14 different augmentation methods using grid search. Despite the simplicity and reduced search effort of our method compared to RandAugment, our approach consistently surpasses RandAugment in performance across various data regimes as shown in \cref{tab_frcnn}.

\begin{table}[t]
\caption{Comparison to baseline method using ResNet-50 Fatser-RCNN on the MS COCO dataset. We report the mean average precision (mAP) metric.}
\centering
\begin{adjustbox}{max width=\linewidth}
\begin{tabular}[b]{l c c c c} 
\toprule 
{\bftab Method} \:\:\:& {\bftab COCO 1\% } \:\:\:& {\bftab COCO 5\% } \:\:\:& {\bftab COCO 10\% }\\
\midrule
Standard Aug. & 9.3 & 19.0  & 22.0  \\
RandAugment \cite{cubuk2020randaugment} & 9.9 & 20.4 & 23.5 \\
Background Aug. & 12.5 & 22.5 & 24.7  \\
\midrule
Soft Teacher~\cite{xu2021end} & 19.5 &  28.8 & 31.1 \\
Soft Teacher + Background Aug. & 22.7 &  30.5 & 32.0 \\
\bottomrule
\end{tabular}
\end{adjustbox}
\label{tab_frcnn}
\end{table}

% Our results using our augmentation policy on the above procedures are shown in Tables \ref{tab_frcnn}. 
Table \ref{tab_frcnn} shows the results using our augmentation policy on the above procedures. The proposed background augmentation policy achieves systematic gains across different amounts of training data used, ranging from +2.7 mAP to +3.5 mAP. 
In comparison, a previous regularization technique applied to ResNet-50 Faster-RCNN achieves a gain ranging from +0.6mAP to +1.5mAP.

\fakeparagraph{Combined with Semi-Supervised Learning. }
We then experiment with semi-supervised learning in object detection. In particular, we select Soft Teacher~\cite{xu2021end} and add our background augmentation data to the \texttt{labeled} dataset, which is used to train the teacher network to annotate the \texttt{unlabeled} dataset for knowledge transfer, which is the remaining data from MS COCO training set. The model (ResNet-50-based Faster-RCNN) is kept the same as in previous experiments. The other training hyper-parameters are based on the MMDetection implementation~\cite{chen2019mmdetection}. 

We compare Soft Teacher with or without our background augmentation as the augmentation strategy. The results are again shown in \cref{tab_frcnn}, from which we can find that our method also effectively improves the mAP.

\subsection{Comparisons on Transformer Architecture}

We also evaluate our background augmentation on Transformer-based architecture. Specifically, we use Swin Transformer small~\cite{liu2021swin} as backbone, Mask-RCNN~\cite{he2017mask, cai2018cascade} as detection framework to train on the full MS COCO dataset. The training hyper-parameters are kept the same as the original configurations~\cite{liu2021swin}, \ie AdamW~\cite{loshchilov2017decoupled} optimizer (initial learning rate of 0.0001, weight decay of 0.05,
and batch size of 16), and 3x schedule (36 epochs with the
learning rate decayed by 10× at epochs 27 and 33). 
The standard augmentation includes horizontal flipping and multi-scale training (resizing the input such that the shorter side is between 480 and 800 while the longer side is at most 1333). 
We report the bounding box AP and instance mask AP metrics following the conventions of transformer evaluation.

\begin{table}[t]
\caption{Comparison to baseline method using Swin Transformer on the MS COCO dataset. We report the bounding box AP and instance mask AP results. }
\centering
\begin{adjustbox}{max width=\linewidth}
\begin{tabular}[b]{l c c c c} 
\toprule 
\multirow{2}{*}{{\bftab Model}} \:\:\:& \multicolumn{2}{c}{{\bftab Swin-T}} \:\:\:& \multicolumn{2}{c}{{\bftab Swin-S}} \\
\cmidrule(lr){2-3}  \cmidrule(lr){4-5} 
& AP$^{\text{box}}$ \:\:\:& AP$^{\text{mask}}$ \:\:\:& AP$^{\text{box}}$ \:\:\:& AP$^{\text{mask}}$ \\
\midrule
Standard Aug. & 46.0  &  41.6 &  48.2 & 43.6 \\
Background Aug. & {\bftab 47.0}  & {\bftab 42.4} & {\bftab 49.0} & {\bftab 44.3}\\
\bottomrule
\end{tabular}
\end{adjustbox}
\label{tab_swin}
\end{table}

\cref{tab_swin} summarizes the results on the Swin transformer. We can find that background augmentation effectively improves the detection and segmentation performance (\eg +1.0 AP$^{\text{box}}$ and +0.8 AP$^{\text{mask}}$ on the Swin-T architecture).

\newcommand{\cm}{\textcolor{Green}{\checkmark}}
\newcommand{\xm}{\textcolor{Red}{$\times$}}

\begin{table}[t]
\caption{Ablation studies on design choices of our background augmentation pipeline.}
\centering
\begin{adjustbox}{max width=\linewidth}
\begin{tabular}[b]{l c c c c c} 
\toprule 
\multirow{2}{*}{{\bftab Model}} \:\:\:& \multirow{2}{*}{\bftab Object Aug.} \:\:\:& \multicolumn{3}{c}{{\bftab Background Aug. }} \:\:\:& {\bftab mAP} \\
  \cmidrule(lr){3-5} 
& &  Prompt & Erosion & Adaptive steps\\
\midrule
RetinaNet &  \xm & \xm & \xm & \xm & 36.7\\
RetinaNet &  \cm & \xm & \xm & \xm & 36.4 \\
\midrule
 & \xm & \xm & \xm & \xm & 35.4 \\
\multirow{2}{*}{RetinaNet}& \xm & \cm & \xm & \xm & 36.8 \\
 & \xm & \cm & \cm & \xm & 37.9 \\
 & \xm & \cm & \cm & \cm & {\bftab 38.3} \\

\bottomrule
\end{tabular}
\end{adjustbox}
\label{tab_ablation}
\end{table}

\subsection{Ablation Studies}

In this section, we conduct ablation studies that verify the design choices of our background augmentation pipeline. In order to do this, we generate different sets of COCO training data and combine them with original data to train a ResNet-50 RetinaNet~\cite{lin2017focal}. We use a similar training schedule and hyper-parameters with previous experiments. 
In particular, we compare several cases: (1) using object augmentation or background augmentation, (2) using background augmentation but with or without our proposed techniques. 
Specifically, for prompt input we either use the COCO caption as prompt or our proposed simple prompt method, and for background mask erosion as well as the adaptive diffusion steps, we choose to either apply them or not. 

Table \ref{tab_ablation} summarizes the results of our ablation experiments, from which we can observe that applying the object augmentation will even decrease the mAP performance (-0.3 mAP). The same situation also applies to background augmentation if the prompt is not chosen correctly. This result indicates that augmenting the data in object detection is non-trivial. With background mask erosion and appropriate prompt choice, we can achieve better detection performance than the baseline (+1.3 mAP on the full COCO dataset). Furthermore, adding the adaptive strength control enhances the final performance to 38.3 mAP. 

\subsection{Augmenting More Data}
In previous experiments, we only generate 1x of COCO data. This is still far from the ability of text-to-image models as it can generate a different image by randomly sampling a different $\mathbf{z}$. In this section, we explore the limits of our background augmentation. Specifically, we experiment with two cases. In the first case, we repeatedly generate $\alpha$ times data for each image and study the effect of increasing $\alpha$ (called uniform sampling).
In the second case, we increase the ratio of augmenting small objects in the training data since detecting small objects often poses more challenges than detecting large objects~\cite{kisantal2019augmentation}. 
Therefore, we increase the proportion of small object detection based on the background area ratio, which is called non-uniform sampling. 
In total, we also generate $\alpha \times$ copy of COCO data but with more augmented data for small objects in this case. 

\begin{figure}[t]
    \centering
    \includegraphics[width=\linewidth]{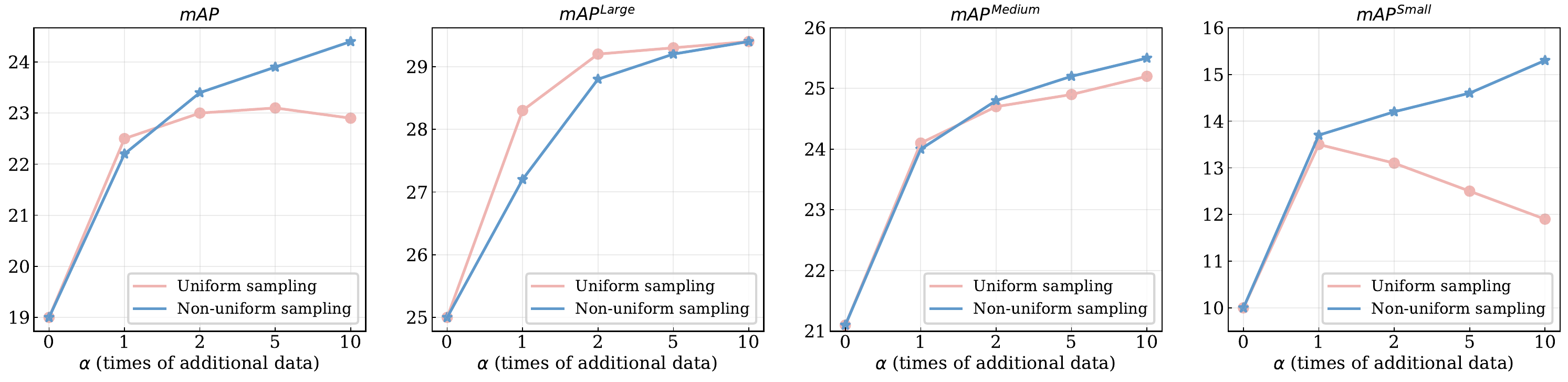}
    \caption{Comparisons of four different mAP metrics on COCO dataset when scaling up the augmented data.}
    \label{fig_mAP}
\end{figure}

We evaluate these two scaling cases in the 10\% COCO data regime. The $\alpha$ is set to 1, 2, 5, and 10 (Note that $\alpha=0$ indicates no background augmentation applied). We use ResNet-50-based Faster RCNN~\cite{ren2015faster} to evaluate different data regimes. 
The training hyper-parameters are kept the same with \cref{sec_odeval}. We report four different metrics: mAP, mAP$^{\text{small}}$, mAP$^{\text{medium}}$, and mAP$^{\text{big}}$, as shown in \cref{fig_mAP}.

From the figure, we find that uniformly increasing the number of augmented data does not consistently increase the mAP performance. 
Observing the results from different sizes of the objects, we find that the major cause is the small object performance, which is decreasing when $\alpha\ge 2$. Instead, non-uniform sampling prioritizes the small objects in augmentation and ensures more diverse images are created in small object detection, leading to better-performed detectors.

\begin{table}[t]
\caption{Comparison to baseline methods using ResNet-50 and Fatser-RCNN on the PASCAL VOC dataset. We report the bounding box mean average precision (mAP$^{\text{box}}$) metric.}
\centering
\begin{adjustbox}{max width=\linewidth}
\begin{tabular}[b]{l c c c c} 
\toprule 
{\bftab Method} \:\:\:& {\bftab RetineNet} \:\:\:& {\bftab Faster-RCNN} \\
\midrule
Standard Aug. & 77.3 & 80.4   \\
RandAugment \cite{cubuk2020randaugment} & 78.2 & 80.9  \\
Background Aug. & {\bftab 79.1} &  {\bftab 81.7} \\
\bottomrule
\end{tabular}
\end{adjustbox}
\label{tab_voc}
\end{table}

\subsection{Evaluation on PASCAL VOC}
To evaluate the effectiveness of our proposed background augmentation on an entirely different dataset and another detection algorithm, we train a Faster R-CNN~\cite{ren2015faster} model and a RetinaNet model~\cite{lin2017focal} with a ResNet-50 backbone on PASCAL VOC dataset~\cite{everingham2010pascal}.
We combine the training sets of PASCAL VOC 2007 and PASCAL VOC 2012, and test our model on the PASCAL VOC
2007 test set (4952 images). Our evaluation metric is the
bounding box mean average precision (mAP$^{\text{box}}$).
The training hyper-parameters are kept the same with previous experiments and the backbone model is initialized from the ImageNet pre-trained checkpoint~\cite{he2016deep}. 
We also include the performance of RandAugment in the PASCAL VOC evaluation. 

The results are summarized in \cref{tab_voc}, where we can find our method leads to a significant increase in terms of box mAP.

\label{sec_pascal}

\section{Conclusion}
In this work, we propose a simple yet effective background augmentation framework for object detection and instance segmentation. We directly utilize the existing mask annotation to perform Inpainting to create new content in the training data. Through a set of rigorous explorations, we confirm that background augmentation, instead of object augmentation, can achieve higher potential. We also propose several techniques to ensure the background contains no additional object so that the generated data can be directly used for augmentation. Results show that our method can be effectively used to improve the detection performance.
Our method has certain limitations: for images containing extremely large objects, background augmentation can hardly alter the content (as shown in \cref{fig_mAP} the improvement of mAP$^{\text{Large}}$ is relatively small). 
In this case, our method could be combined with an effective object augmentation (that may require finetuning of the text-to-image model) to enrich the diversity of this type of data.

\section*{Acknowledgement}
This work is sponsored by Sony AI.

\bibliographystyle{splncs04}
\bibliography{main}

\begin{thebibliography}{10}
\providecommand{\url}[1]{\texttt{#1}}
\providecommand{\urlprefix}{URL }
\providecommand{\doi}[1]{https://doi.org/#1}

\bibitem{azizi2023synthetic}
Azizi, S., Kornblith, S., Saharia, C., Norouzi, M., Fleet, D.J.: Synthetic data from diffusion models improves imagenet classification. arXiv preprint arXiv:2304.08466  (2023)

\bibitem{besnier2020dataset}
Besnier, V., Jain, H., Bursuc, A., Cord, M., P{\'e}rez, P.: This dataset does not exist: training models from generated images. In: ICASSP 2020-2020 IEEE International Conference on Acoustics, Speech and Signal Processing (ICASSP). pp.~1--5. IEEE (2020)

\bibitem{cai2018cascade}
Cai, Z., Vasconcelos, N.: Cascade r-cnn: Delving into high quality object detection. In: Proceedings of the IEEE conference on computer vision and pattern recognition. pp. 6154--6162 (2018)

\bibitem{chen2019mmdetection}
Chen, K., Wang, J., Pang, J., Cao, Y., Xiong, Y., Li, X., Sun, S., Feng, W., Liu, Z., Xu, J., et~al.: Mmdetection: Open mmlab detection toolbox and benchmark. arXiv preprint arXiv:1906.07155  (2019)

\bibitem{cordts2016cityscapes}
Cordts, M., Omran, M., Ramos, S., Rehfeld, T., Enzweiler, M., Benenson, R., Franke, U., Roth, S., Schiele, B.: The cityscapes dataset for semantic urban scene understanding. In: Proceedings of the IEEE conference on computer vision and pattern recognition. pp. 3213--3223 (2016)

\bibitem{creswell2018generative}
Creswell, A., White, T., Dumoulin, V., Arulkumaran, K., Sengupta, B., Bharath, A.A.: Generative adversarial networks: An overview. IEEE signal processing magazine  \textbf{35}(1),  53--65 (2018)

\bibitem{cubuk2018autoaugment}
Cubuk, E.D., Zoph, B., Mane, D., Vasudevan, V., Le, Q.V.: Autoaugment: Learning augmentation policies from data. arXiv preprint arXiv:1805.09501  (2018)

\bibitem{cubuk2020randaugment}
Cubuk, E.D., Zoph, B., Shlens, J., Le, Q.V.: Randaugment: Practical automated data augmentation with a reduced search space. In: Proceedings of the IEEE/CVF conference on computer vision and pattern recognition workshops. pp. 702--703 (2020)

\bibitem{dhariwal2021diffusion}
Dhariwal, P., Nichol, A.: Diffusion models beat gans on image synthesis. Advances in neural information processing systems  \textbf{34},  8780--8794 (2021)

\bibitem{vit}
Dosovitskiy, A., Beyer, L., Kolesnikov, A., Weissenborn, D., Zhai, X., Unterthiner, T., Dehghani, M., Minderer, M., Heigold, G., Gelly, S., et~al.: An image is worth 16x16 words: Transformers for image recognition at scale. arXiv preprint arXiv:2010.11929  (2020)

\bibitem{dosovitskiy2015flownet}
Dosovitskiy, A., Fischer, P., Ilg, E., Hausser, P., Hazirbas, C., Golkov, V., Van Der~Smagt, P., Cremers, D., Brox, T.: Flownet: Learning optical flow with convolutional networks. In: Proceedings of the IEEE international conference on computer vision. pp. 2758--2766 (2015)

\bibitem{everingham2010pascal}
Everingham, M., Van~Gool, L., Williams, C.K., Winn, J., Zisserman, A.: The pascal visual object classes (voc) challenge. International journal of computer vision  \textbf{88},  303--338 (2010)

\bibitem{feng2021tood}
Feng, C., Zhong, Y., Gao, Y., Scott, M.R., Huang, W.: Tood: Task-aligned one-stage object detection. In: 2021 IEEE/CVF International Conference on Computer Vision (ICCV). pp. 3490--3499. IEEE Computer Society (2021)

\bibitem{feng2022promptdet}
Feng, C., Zhong, Y., Jie, Z., Chu, X., Ren, H., Wei, X., Xie, W., Ma, L.: Promptdet: Towards open-vocabulary detection using uncurated images. In: European Conference on Computer Vision. pp. 701--717. Springer (2022)

\bibitem{ganin2016domain}
Ganin, Y., Ustinova, E., Ajakan, H., Germain, P., Larochelle, H., Laviolette, F., Marchand, M., Lempitsky, V.: Domain-adversarial training of neural networks. The journal of machine learning research  \textbf{17}(1),  2096--2030 (2016)

\bibitem{hafiz2020survey}
Hafiz, A.M., Bhat, G.M.: A survey on instance segmentation: state of the art. International journal of multimedia information retrieval  \textbf{9}(3),  171--189 (2020)

\bibitem{he2021masked}
He, K., Chen, X., Xie, S., Li, Y., Doll’ar, P., Girshick, R.B.: Masked autoencoders are scalable vision learners. 2022 ieee. In: CVF Conference on Computer Vision and Pattern Recognition (CVPR). pp. 15979--15988 (2021)

\bibitem{he2017mask}
He, K., Gkioxari, G., Doll{\'a}r, P., Girshick, R.: Mask r-cnn. In: Proceedings of the IEEE international conference on computer vision. pp. 2961--2969 (2017)

\bibitem{he2016deep}
He, K., Zhang, X., Ren, S., Sun, J.: Deep residual learning for image recognition. In: Proceedings of the IEEE conference on computer vision and pattern recognition. pp. 770--778 (2016)

\bibitem{he2022synthetic}
He, R., Sun, S., Yu, X., Xue, C., Zhang, W., Torr, P., Bai, S., Qi, X.: Is synthetic data from generative models ready for image recognition? arXiv preprint arXiv:2210.07574  (2022)

\bibitem{hnewa2020object}
Hnewa, M., Radha, H.: Object detection under rainy conditions for autonomous vehicles: A review of state-of-the-art and emerging techniques. IEEE Signal Processing Magazine  \textbf{38}(1),  53--67 (2020)

\bibitem{ho2022classifier}
Ho, J., Salimans, T.: Classifier-free diffusion guidance. arXiv preprint arXiv:2207.12598  (2022)

\bibitem{jahanian2021generative}
Jahanian, A., Puig, X., Tian, Y., Isola, P.: Generative models as a data source for multiview representation learning. arXiv preprint arXiv:2106.05258  (2021)

\bibitem{kisantal2019augmentation}
Kisantal, M., Wojna, Z., Murawski, J., Naruniec, J., Cho, K.: Augmentation for small object detection. arXiv preprint arXiv:1902.07296  (2019)

\bibitem{li2022exploring}
Li, Y., Mao, H., Girshick, R., He, K.: Exploring plain vision transformer backbones for object detection. In: European Conference on Computer Vision. pp. 280--296. Springer (2022)

\bibitem{li2024synthetic}
Li, Y., Dong, X., Chen, C., Li, J., Wen, Y., Spranger, M., Lyu, L.: Is synthetic image useful for transfer learning? an investigation into data generation, volume, and utilization. arXiv preprint arXiv:2403.19866  (2024)

\bibitem{lim2019fastautoaug}
Lim, S., Kim, I., Kim, T., Kim, C., Kim, S.: Fast autoaugment. Advances in Neural Information Processing Systems  \textbf{32},  6665--6675 (2019)

\bibitem{lin2017feature}
Lin, T.Y., Doll{\'a}r, P., Girshick, R., He, K., Hariharan, B., Belongie, S.: Feature pyramid networks for object detection. In: Proceedings of the IEEE conference on computer vision and pattern recognition. pp. 2117--2125 (2017)

\bibitem{lin2017focal}
Lin, T.Y., Goyal, P., Girshick, R., He, K., Doll{\'a}r, P.: Focal loss for dense object detection. In: Proceedings of the IEEE international conference on computer vision. pp. 2980--2988 (2017)

\bibitem{lin2014microsoft}
Lin, T.Y., Maire, M., Belongie, S., Hays, J., Perona, P., Ramanan, D., Doll{\'a}r, P., Zitnick, C.L.: Microsoft coco: Common objects in context. In: Computer Vision--ECCV 2014: 13th European Conference, Zurich, Switzerland, September 6-12, 2014, Proceedings, Part V 13. pp. 740--755. Springer (2014)

\bibitem{litjens2017survey}
Litjens, G., Kooi, T., Bejnordi, B.E., Setio, A.A.A., Ciompi, F., Ghafoorian, M., Van Der~Laak, J.A., Van~Ginneken, B., S{\'a}nchez, C.I.: A survey on deep learning in medical image analysis. Medical image analysis  \textbf{42},  60--88 (2017)

\bibitem{liu2021swin}
Liu, Z., Lin, Y., Cao, Y., Hu, H., Wei, Y., Zhang, Z., Lin, S., Guo, B.: Swin transformer: Hierarchical vision transformer using shifted windows. In: Proceedings of the IEEE/CVF international conference on computer vision. pp. 10012--10022 (2021)

\bibitem{loshchilov2017decoupled}
Loshchilov, I., Hutter, F.: Decoupled weight decay regularization. arXiv preprint arXiv:1711.05101  (2017)

\bibitem{lu2022dpm}
Lu, C., Zhou, Y., Bao, F., Chen, J., Li, C., Zhu, J.: Dpm-solver: A fast ode solver for diffusion probabilistic model sampling in around 10 steps. Advances in Neural Information Processing Systems  \textbf{35},  5775--5787 (2022)

\bibitem{lu2022dpm++}
Lu, C., Zhou, Y., Bao, F., Chen, J., Li, C., Zhu, J.: Dpm-solver++: Fast solver for guided sampling of diffusion probabilistic models. arXiv preprint arXiv:2211.01095  (2022)

\bibitem{lugmayr2022repaint}
Lugmayr, A., Danelljan, M., Romero, A., Yu, F., Timofte, R., Van~Gool, L.: Repaint: Inpainting using denoising diffusion probabilistic models. In: Proceedings of the IEEE/CVF Conference on Computer Vision and Pattern Recognition. pp. 11461--11471 (2022)

\bibitem{nichol2021glide}
Nichol, A., Dhariwal, P., Ramesh, A., Shyam, P., Mishkin, P., McGrew, B., Sutskever, I., Chen, M.: Glide: Towards photorealistic image generation and editing with text-guided diffusion models. arXiv preprint arXiv:2112.10741  (2021)

\bibitem{peng2017visda}
Peng, X., Usman, B., Kaushik, N., Hoffman, J., Wang, D., Saenko, K.: Visda: The visual domain adaptation challenge. arXiv preprint arXiv:1710.06924  (2017)

\bibitem{radford2021learning}
Radford, A., Kim, J.W., Hallacy, C., Ramesh, A., Goh, G., Agarwal, S., Sastry, G., Askell, A., Mishkin, P., Clark, J., et~al.: Learning transferable visual models from natural language supervision. In: International conference on machine learning. pp. 8748--8763. PMLR (2021)

\bibitem{ramesh2022hierarchical}
Ramesh, A., Dhariwal, P., Nichol, A., Chu, C., Chen, M.: Hierarchical text-conditional image generation with clip latents. arXiv preprint arXiv:2204.06125  \textbf{1}(2), ~3 (2022)

\bibitem{redmon2016you}
Redmon, J., Divvala, S., Girshick, R., Farhadi, A.: You only look once: Unified, real-time object detection. In: Proceedings of the IEEE conference on computer vision and pattern recognition. pp. 779--788 (2016)

\bibitem{ren2015faster}
Ren, S., He, K., Girshick, R., Sun, J.: Faster r-cnn: Towards real-time object detection with region proposal networks. Advances in neural information processing systems  \textbf{28} (2015)

\bibitem{richter2016playing}
Richter, S.R., Vineet, V., Roth, S., Koltun, V.: Playing for data: Ground truth from computer games. In: Computer Vision--ECCV 2016: 14th European Conference, Amsterdam, The Netherlands, October 11-14, 2016, Proceedings, Part II 14. pp. 102--118. Springer (2016)

\bibitem{rombach2022high}
Rombach, R., Blattmann, A., Lorenz, D., Esser, P., Ommer, B.: High-resolution image synthesis with latent diffusion models. In: Proceedings of the IEEE/CVF conference on computer vision and pattern recognition. pp. 10684--10695 (2022)

\bibitem{ronneberger2015u}
Ronneberger, O., Fischer, P., Brox, T.: U-net: Convolutional networks for biomedical image segmentation. In: International Conference on Medical image computing and computer-assisted intervention. pp. 234--241. Springer (2015)

\bibitem{saharia2022palette}
Saharia, C., Chan, W., Chang, H., Lee, C., Ho, J., Salimans, T., Fleet, D., Norouzi, M.: Palette: Image-to-image diffusion models. In: ACM SIGGRAPH 2022 Conference Proceedings. pp. 1--10 (2022)

\bibitem{saharia2022photorealistic}
Saharia, C., Chan, W., Saxena, S., Li, L., Whang, J., Denton, E.L., Ghasemipour, K., Gontijo~Lopes, R., Karagol~Ayan, B., Salimans, T., et~al.: Photorealistic text-to-image diffusion models with deep language understanding. Advances in Neural Information Processing Systems  \textbf{35},  36479--36494 (2022)

\bibitem{shin2023fill}
Shin, J., Kang, M., Park, J.: Fill-up: Balancing long-tailed data with generative models. arXiv preprint arXiv:2306.07200  (2023)

\bibitem{shorten2019survey}
Shorten, C., Khoshgoftaar, T.M.: A survey on image data augmentation for deep learning. Journal of Big Data  \textbf{6}(1),  1--48 (2019)

\bibitem{tan2020efficientdet}
Tan, M., Pang, R., Le, Q.V.: Efficientdet: Scalable and efficient object detection. In: Proceedings of the IEEE/CVF conference on computer vision and pattern recognition. pp. 10781--10790 (2020)

\bibitem{tian2023stablerep}
Tian, Y., Fan, L., Isola, P., Chang, H., Krishnan, D.: Stablerep: Synthetic images from text-to-image models make strong visual representation learners. arXiv preprint arXiv:2306.00984  (2023)

\bibitem{tramer2017ensemble}
Tram{\`e}r, F., Kurakin, A., Papernot, N., Goodfellow, I., Boneh, D., McDaniel, P.: Ensemble adversarial training: Attacks and defenses. arXiv preprint arXiv:1705.07204  (2017)

\bibitem{weng2023diffusion}
Weng, Z., Bravo-S{\'a}nchez, L., Yeung, S.: Diffusion-hpc: Generating synthetic images with realistic humans. arXiv preprint arXiv:2303.09541  (2023)

\bibitem{wu2024datasetdm}
Wu, W., Zhao, Y., Chen, H., Gu, Y., Zhao, R., He, Y., Zhou, H., Shou, M.Z., Shen, C.: Datasetdm: Synthesizing data with perception annotations using diffusion models. Advances in Neural Information Processing Systems  \textbf{36} (2024)

\bibitem{wu2023diffumask}
Wu, W., Zhao, Y., Shou, M.Z., Zhou, H., Shen, C.: Diffumask: Synthesizing images with pixel-level annotations for semantic segmentation using diffusion models. arXiv preprint arXiv:2303.11681  (2023)

\bibitem{xie2023smartbrush}
Xie, S., Zhang, Z., Lin, Z., Hinz, T., Zhang, K.: Smartbrush: Text and shape guided object inpainting with diffusion model. In: Proceedings of the IEEE/CVF Conference on Computer Vision and Pattern Recognition. pp. 22428--22437 (2023)

\bibitem{xu2021end}
Xu, M., Zhang, Z., Hu, H., Wang, J., Wang, L., Wei, F., Bai, X., Liu, Z.: End-to-end semi-supervised object detection with soft teacher. In: Proceedings of the IEEE/CVF International Conference on Computer Vision. pp. 3060--3069 (2021)

\bibitem{xue2023freestyle}
Xue, H., Huang, Z., Sun, Q., Song, L., Zhang, W.: Freestyle layout-to-image synthesis. In: Proceedings of the IEEE/CVF Conference on Computer Vision and Pattern Recognition. pp. 14256--14266 (2023)

\bibitem{yang2024freemask}
Yang, L., Xu, X., Kang, B., Shi, Y., Zhao, H.: Freemask: Synthetic images with dense annotations make stronger segmentation models. Advances in Neural Information Processing Systems  \textbf{36} (2024)

\bibitem{yang2021artificial}
Yang, R., Yu, Y.: Artificial convolutional neural network in object detection and semantic segmentation for medical imaging analysis. Frontiers in oncology  \textbf{11},  638182 (2021)

\bibitem{yun2019cutmix}
Yun, S., Han, D., Oh, S.J., Chun, S., Choe, J., Yoo, Y.: Cutmix: Regularization strategy to train strong classifiers with localizable features. In: Proceedings of the IEEE/CVF international conference on computer vision. pp. 6023--6032 (2019)

\bibitem{yun2020videomix}
Yun, S., Oh, S.J., Heo, B., Han, D., Kim, J.: Videomix: Rethinking data augmentation for video classification. arXiv preprint arXiv:2012.03457  (2020)

\bibitem{zhang2021datasetgan}
Zhang, Y., Ling, H., Gao, J., Yin, K., Lafleche, J.F., Barriuso, A., Torralba, A., Fidler, S.: Datasetgan: Efficient labeled data factory with minimal human effort. In: Proceedings of the IEEE/CVF Conference on Computer Vision and Pattern Recognition. pp. 10145--10155 (2021)

\bibitem{zoph2020learning}
Zoph, B., Cubuk, E.D., Ghiasi, G., Lin, T.Y., Shlens, J., Le, Q.V.: Learning data augmentation strategies for object detection. In: Computer Vision--ECCV 2020: 16th European Conference, Glasgow, UK, August 23--28, 2020, Proceedings, Part XXVII 16. pp. 566--583. Springer (2020)

\end{thebibliography}
\end{document}